\documentclass[sigconf]{acmart}
\AtBeginDocument{%
  }

\setcopyright{acmlicensed}
\copyrightyear{2024}
\acmYear{2024}
\acmDOI{XXXXXXX.XXXXXXX}

\acmConference[WWW '25]{The Web Conference}{28 April - 2 May 2025}{Sydney, Australia}
\acmISBN{978-1-4503-XXXX-X/18/06}

\begin{document}

\title{Efficient user history modeling with amortized inference for deep learning recommendation models}

\author{Lars Hertel}
\orcid{1234-5678-9012}
\affiliation{%
  \institution{LinkedIn Corporation}
  \city{Sunnyvale}
  \state{California}
  \country{USA}
}

\author{Neil Daftary}

\affiliation{%
  \institution{LinkedIn Corporation}
  \city{Sunnyvale}
  \state{California}
  \country{USA}
}

\author{Fedor Borisyuk}

\affiliation{%
  \institution{LinkedIn Corporation}
  \city{Sunnyvale}
  \state{California}
  \country{USA}
}

\author{Aman Gupta}
\affiliation{%
  \institution{LinkedIn Corporation}
  \city{Sunnyvale}
  \state{California}
  \country{USA}
}

\author{Rahul Mazumder}
\affiliation{%
  \institution{LinkedIn Corporation}
  \city{Sunnyvale}
  \state{California}
  \country{USA}
}


\renewcommand{\shortauthors}{Hertel et al.}

\begin{abstract}
We study user history modeling via Transformer encoders in deep learning recommendation models (DLRM). Such architectures can significantly improve recommendation quality, but usually incur high latency cost necessitating infrastructure upgrades or very small Transformer models. An important part of user history modeling is early fusion of the candidate item and various methods have been studied. We revisit early fusion and compare concatenation of the candidate to each history item against appending it to the end of the list as a separate item. Using the latter method, allows us to reformulate the recently proposed amortized history inference algorithm M-FALCON \cite{zhai2024actions} for the case of DLRM models. We show via experimental results that appending with cross-attention performs on par with concatenation and that amortization significantly reduces inference costs. We conclude with results from deploying this model on the LinkedIn Feed and Ads surfaces, where amortization reduces latency by 30\% compared to non-amortized inference.
\end{abstract}




\maketitle

\section{Introduction}
User interaction history plays a crucial role in deep learning recommendation models (DLRM). Items that a user interacted with can be encoded with embeddings and mean-pooled. However, more recently simple pooling has been replaced with pairwise attention via Deep Interest Networks (DIN) \cite{zhou2018deep} and with Transformers in Behavioral Sequence Transformers (BST) \cite{chen2019behavior} and TransAct \cite{xia2023transact}. In particular, BST and TransAct differ in their methods of early fusion. Early fusion is the concept of integrating a candidate item early on in the ranking process to be able to extract relevant signals from the user history. 
A major challenge of Transformer-based user history models is the online serving cost. According to \cite{xia2023transact}, TransAct increased computational complexity by 65 times compared to the baseline, resulting in a 24x latency increase on CPU \cite{GPUacceleratedInference}. The authors thus go on to describe how they migrated their system to be served on GPUs. Even smaller architectures such as DIN or BST can show increases of 25\% and 53\% in latency \cite{chen2019behavior}, respectively.
In this study, we propose to revisit the choice of early fusion with the goal of leveraging amortized history inference similar to the M-FALCON algorithm \cite{zhai2024actions} for generative recommenders.

Specifically, we study two methods of early fusion: concatenating the candidate item to each history step or appending it to the end of the list. For the latter method we formulate an amortized inference version that significantly reduces the number of computations. We demonstrate via experimental results on public datasets and internal Feed and Ads ranking systems that concatenating and appending perform comparably in terms of engagement prediction. In addition we visualize the attention matrices of both early fusion approaches which result in very different patterns, indicating that the two approaches learn different models. Finally, we show through benchmarks and real world deployment how amortized inference can reduce the latency cost of Transformer based user history modeling.

\section{Methods}

We focus on DLRM-style recommender systems. This means pointwise ranking where each ranked item is a separate input to the model. Given a feature vector containing user and item information, an MLP, optionally including a feature interaction module, transforms the features into predictions of actions the user may take on the item. Our focus is on encoding the sequence of items that the user engaged with in the past as an input feature to the MLP. Let the sequence of engaged items be represented by $H_0, \ldots, H_n$. Here $H_i$ represents the sequence embedding features of dimension $d$ corresponding to the $i$-th interacted item. Furthermore, let the corresponding features of the candidate item that is currently being ranked be $C$, also of dimension $d$.

\subsection{Early Fusion: Appending vs. Concatenating}
We review two methods of early fusion, namely, appending the candidate item to the sequence (append) and concatenating the candidate item to each sequence item (concat).\\
\textbf{Append}: BST \cite{chen2019behavior} encodes the user history by appending the candidate item to the history and transforming by a Transformer-Encoder, that is, 
\begin{equation}
\label{eq:bst}
\mathrm{Transformer}([H_1, \ldots, H_n, C]).
\end{equation}
\textbf{Concat}: TransAct \cite{xia2023transact} on the other hand concatenates the candidate to each interacted item:
\begin{equation}
\mathrm{Transformer}([(H_1, C), \ldots, (H_n, C)]).
\end{equation}
Specifically, the authors mention that concatenating performs better in offline results on their use case.\\
We propose to append the candidate item, but using cross-attention so that history items cannot attend to the candidate. Therefore we define,

\begin{align}\label{eq:cross-att}
Q & = W_q [H_1, \ldots, H_n, C] \\
K & = W_k [H_1, \ldots, H_n] \\
V & = W_v [H_1, \ldots, H_n]
\end{align}
and apply a Transformer on these. For simplicity, we will refer to this method of appending with cross-attention from here on when we refer to appending. We note that under the same hyperparameters appending and concatenating lead to a different number of parameters due to the different input sizes to the Transformer. In order to match the number of parameters between the two methods, we tune the key dimension and the feedforward dimension of the Transformer. The key dimension is the projection dimension of $W_q$ and $W_k$. For simplicity, we also use the same dimensions for $W_v$. The feedforward dimension is the projection dimension of the feedforward network of the Transformer. We further note that we mask padding during multihead attention.


\subsection{Amortized Inference for User Interaction History Encoders}\label{sec:methods-amortized}
\begin{figure}
  \centering
  \includegraphics[width=\linewidth]{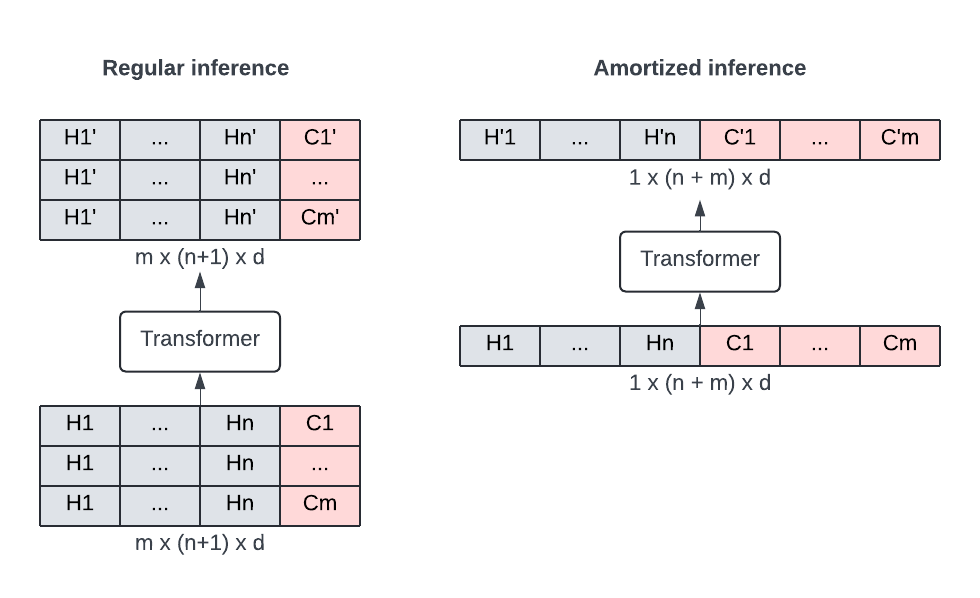}
  \caption{Illustration of regular inference (left) for user action history architectures vs. amortized inference (right). History items are shown in grey and candidate items in red. In amortized inference candidate items are added to the sequence causing the Transformer to only process one sample per request instead of $m$ samples.}
  \label{fig:amortized}
  \Description{A figure showing two diagrams. First for regular inference a three by four grid where history sequences are combined with candidates in the last column. This goes into a Transformer. The output is of the same shape. The amortized inference figure shows a grid of one row and six columns, where the left three are history items and the right three are candidate items. This goes into a Transformer and the output is of the same shape. The figure indicates the input and output shapes under regular inference compared to amortized inference. For amortized inference the input and output is smaller.}
\end{figure}

As part of generative recommendation models, \cite{zhai2024actions} recently proposed the M-FALCON algorithm to accelerate inference. Figure~\ref{fig:amortized} shows regular inference compared to amortized inference. In our case, during online inference ranking models score $m$ candidate items for a user. In DLRM-type models, each candidate commonly constitutes an input example as in Figure~\ref{fig:amortized} (left). However, for user action history modeling the user history is constant across all candidates and computations associated with the history are repeated. Amortized inference as shown in Figure~\ref{fig:amortized} (right) proposes to instead append all $m$ candidates to the sequence \cite{zhai2024actions}. Under cross-attention described in Equation \eqref{eq:cross-att}, the candidate outputs $[C'_1, \ldots, C'_m]$ are equivalent to those from regular inference. However, other components in a DLRM-style model such as the MLP are not directly compatible with this inference format. We therefore apply appropriate reshaping from $m \times (n+1) \times d$ to $1 \times (n+m) \times d$ before the Transformer, and reshaping back to the regular format from $1 \times (n+m) \times d$ to $m \times d$ candidate outputs after the Transformer. This makes amortized inference for the Transformer compatible with the other model components in a DLRM such as the MLP. 

\section{Results}


\subsection{Can appending match the performance of concatenating?}
We introduced user action history encoding where we append the candidate and use cross-attention in Equation \eqref{eq:cross-att}. Cross-attention is required to leverage amortized inference as described in Section~\ref{sec:methods-amortized}. However, we first would like to establish that appending with cross-attention works as well as concatenating in terms of prediction accuracy. To this end, we compare the two methods on four public and two internal datasets.
\begin{itemize}
    \item \textbf{MovieLens 20M}: 20 million movie ratings collected from Movielens.com.
    \item \textbf{Amazon Books}: Ratings for books on Amazon.com.
    \item \textbf{Goodbooks}: Ratings for the 10,000 most popular books on Goodreads.com.
    \item \textbf{Netflix}: Subset of the Netflix Prize competition dataset. Contains ratings for movies on Netflix.
    \item \textbf{LinkedIn Feed}: Ranking model trained to predict contributions (like / comment / share). We provide offline engagement improvement over a baseline without user action history encoding. A difference of 0.05\% is considered relevant.
    \item \textbf{LinkedIn Ads}: Ranking model trained to predict clicks. Provided is the AUC improvement over a baseline without user history modeling. An improvement of 0.1\% is considered relevant \cite{cheng2016wide, zhou2018deep}.
\end{itemize}
For the public datasets we create sequences of user ratings and predict the last rating given an embedding for the item that is being rated. Where timestamps are available, we split the data into training, validation, and test data using the 80th, 90th, and 100th percentile of timestamps \cite{sun2023take}. For the rating prediction we add an MLP on top of the sequence encoder output. If available the MLP incorporates the user ID embedding. The model is trained with a mean squared error loss and evaluated with the mean absolute error (MAE) on the test data. The Transformer hyperparameters for public and internal datasets are provided in Table~\ref{tab:hyperparameters}.

\begin{table}[]
  \caption{Hyperparameter settings for each method and dataset combination.}
  \label{tab:hyperparameters}
\begin{tabular}{@{}lllllll@{}}
\toprule
                & Method & \begin{tabular}[c]{@{}l@{}}Emb\\ dim\end{tabular} & \begin{tabular}[c]{@{}l@{}}Ffwd / \\ Key dim\end{tabular} & \begin{tabular}[c]{@{}l@{}}Num\\ layers\end{tabular} & \begin{tabular}[c]{@{}l@{}}Num\\ heads\end{tabular} & \begin{tabular}[c]{@{}l@{}}Seq\\ length\end{tabular} \\ \midrule
\textit{Public} & Append & 16                                                & 24                                                        & 2                                                    & 1                                                   & 50                                                   \\
\textit{}       & Concat & 16                                                & 16                                                        & 2                                                    & 1                                                   & 50                                                   \\
\textit{Feed}   & Append & 54                                                & 40                                                        & 2                                                    & 1                                                   & 48                                                   \\
                & Concat & 104                                               & 24                                                        & 2                                                    & 1                                                   & 48                                                   \\
\textit{Ads}    & Append & 24                                                & 32                                                        & 1                                                    & 4                                                   & 20                                                   \\
                & Concat & 40                                                & 16                                                        & 1                                                    & 4                                                   & 20                                                   \\ \bottomrule
\end{tabular}
\end{table}

Table~\ref{tab:benchmark-results} shows results for each dataset using appending and concatenating. Bold face marks the better performing method for each dataset.

\begin{table}
  \caption{Evaluation results of comparing appending for early fusion vs. concatenating.}
  \label{tab:benchmark-results}
  \begin{tabular}{ccc}
    \toprule
    Dataset       & Append & Concat \\ \midrule
    MovieLens 20M (MAE $\downarrow$) & \textbf{0.709} & 0.724          \\
    Amazon Books (MAE $\downarrow$)  & \textbf{0.622} & 0.681          \\
    Goodbooks (MAE $\downarrow$)     & 0.722          & \textbf{0.71}  \\
    Netflix (MAE $\downarrow$)      & 0.727          & \textbf{0.722} \\ 
    Feed (offline engagement $\uparrow$) & \textbf{+0.18\%}          & \textbf{+0.16\%} \\
    Ads (offline AUC $\uparrow$) & \textbf{+0.8130\%}          & \textbf{+0.8125\%} \\\bottomrule
\end{tabular}
\end{table}

We observe that either method performs better on two out of four benchmark datasets. On the LinkedIn Feed and Ads ranking models the differences in performance are within significance thresholds. We conclude that appending performs similar to concatenating and can be used in place of it to leverage amortized inference.

\subsection{What does concatenating learn vs. what does appending learn?}
\begin{figure}
  \centering
  \includegraphics[width=.7\linewidth]{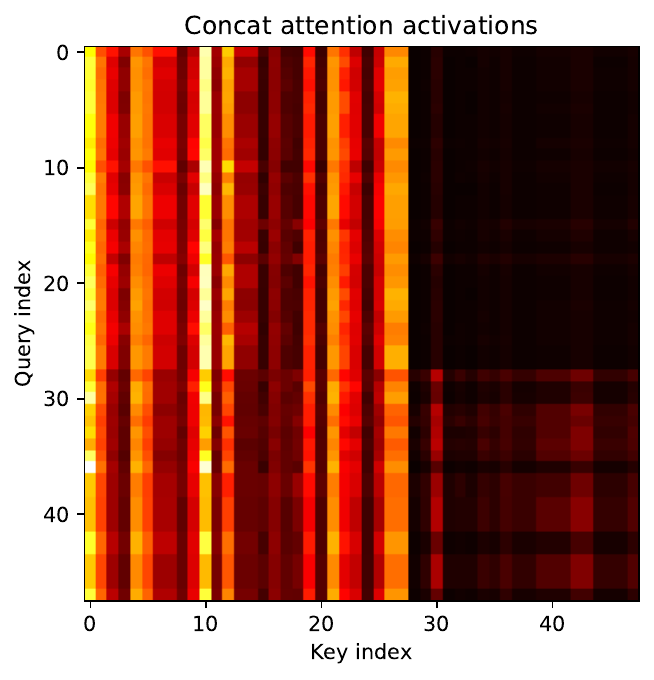}
  \includegraphics[width=.7\linewidth]{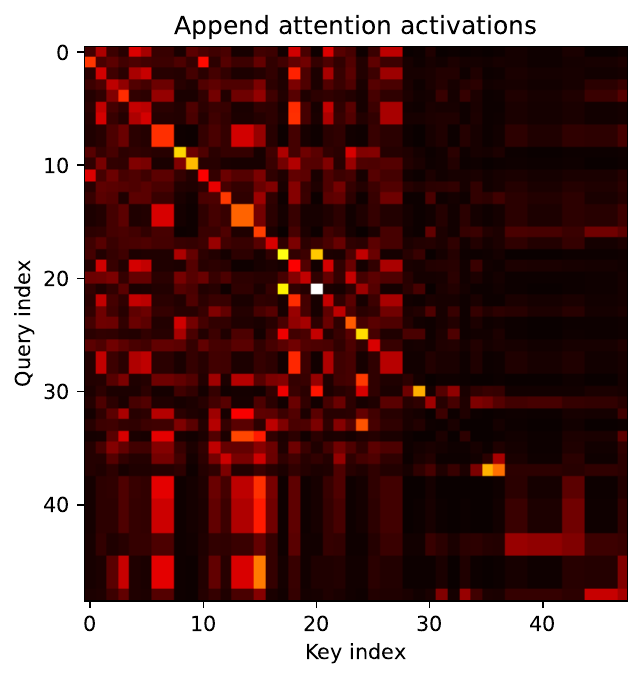}
  \caption{Attention activations from the first Transformer layer for early fusion with concatenating (top) and appending with cross-attention (bottom) for a positive Feed training example.}
  \label{fig:attention}
  \Description{Two visualizations that show attention activations for concatenating and appending for early fusion. Query dimension is on the vertical axis, key dimension on the horizontal axis. Brightness indicates the magnitude of the activation value. The figure for concatenating looks like vertical stripes of different colors. This means that many of the query indices (vertical axis) have the same attention patterns across key indices (horizontal axis). The figure for appending looks dark with squares of different sizes being slightly lighter. There is no clear pattern in the append figure except for brightness along the vertical axis. This indicates that every query attends to different keys and all items attend somewhat to themselves.}
\end{figure}

We want to further understand what each early fusion method is modeling. To do this, we analyze attention matrices for each method from the first Transformer layer for an example from the training data of the Feed dataset as shown in Figure~\ref{fig:attention}. We notice that in the case of concatenating (Figure~\ref{fig:attention} top), values are nearly constant across the query index dimension. We believe that since each step in the sequence contains the candidate, the model may learn pairwise attention between the historical item and the candidate similar to DIN \cite{zhou2018deep}. In this case the model does not necessarily need to communicate information across sequence steps. After inspection of the activations for Q and K of this example (not shown), we find that those activations are all close to zero except for one dimension. This indicates that there may be room for parameter reduction in the concatenation approach in the case of our Feed task. The attention activations for appending (Figure~\ref{fig:attention} bottom) show different attention patterns for every query index with a diagonal indicating items attending to themselves. This indicates that information is propagated across sequence steps.

\subsection{When does amortized inference provide maximum improvements?}\label{sec:benchmarking}
\begin{figure}
  \centering
  \includegraphics[width=\linewidth]{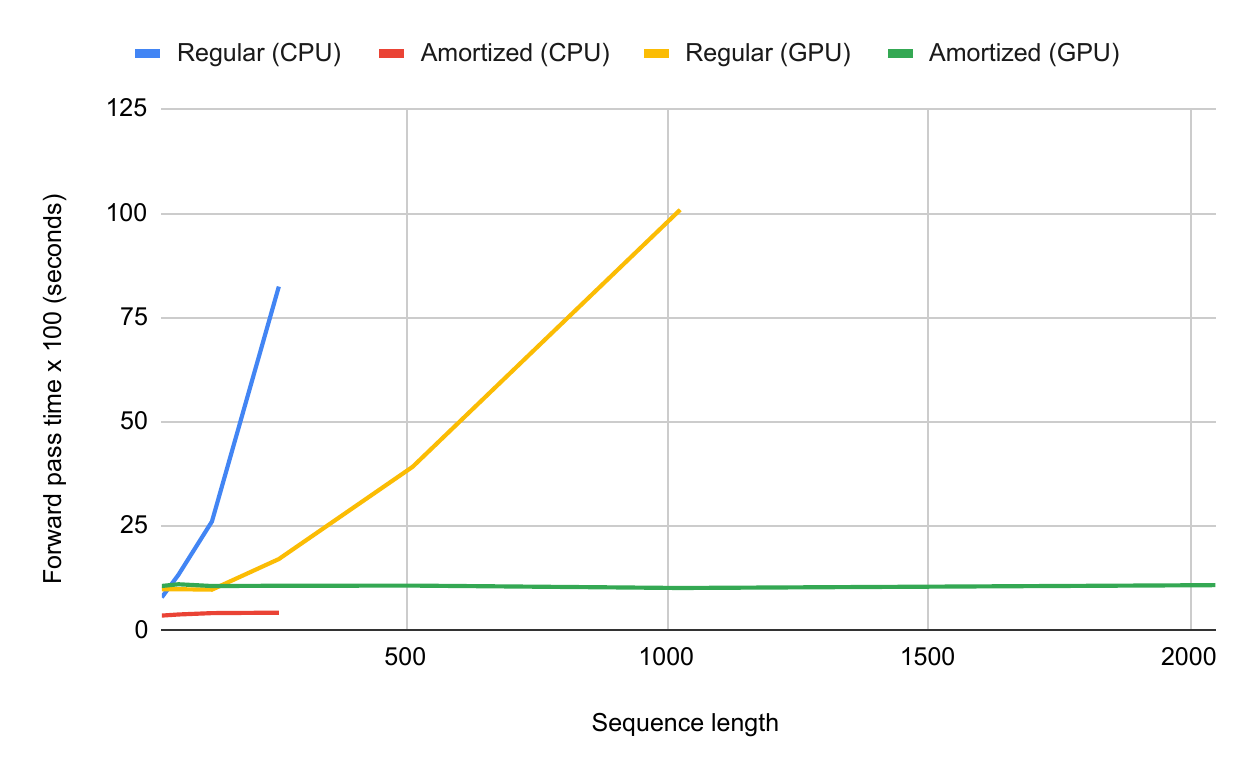}
  \caption{Inference time for 100 forward passes using regular vs. amortized inference on CPU and GPU.}
  \label{fig:time-sequence-length}
  \Description{A graph with forward pass time x 100 (seconds) on the vertical axis and sequence length on the horizontal axis. Four lines show the inference time for the combinations: regular inference on CPU, amortized inference on CPU, regular inference on GPU, amortized inference on GPU. The line for regular inference on CPU seems to shoot up compared to the amortized-CPU line. Both lines only go up to 256 sequence length on the horizontal axis. For GPU, the regular line grows quickly while the amortized line seems to stay constant.}
\end{figure}

We want to further investigate amortized inference to understand when it provides maximal benefits. For this, we first consider the theoretical complexity of regular vs. amortized inference in terms of the number of floating point operations (FLOPS). Regular inference has complexity $O(lmnd^2+lmn^2d)$. Here, $l$ is the number of Transformer layers, $n$ is the history length, $m$ the number of candidates, and $d$ the embedding dimension. The first term corresponds to projections in multi-head attention and the feedforward network and the second term to dot-product attention. The theoretical complexity of amortized inference is $O(l(n+m)d^2+l(n+m)^2d)$.  We can see that $n+m < nm$ for any numbers larger than two. Furthermore, while the ratio of the two is constant as $l$ grows, it increases linearly as $n$ grows. In other words, the benefits of amortized inference increase as the sequence length grows. These observations are confirmed via benchmarking across 100 forward passes as shown in Figure~\ref{fig:time-sequence-length} for varying sequence lengths. In the CPU setting we use the Feed Transformer hyperparameters in Table~\ref{tab:hyperparameters} and $m=512$. For the GPU benchmark we increase the model size to $l=8$, $d=512$, and key/feedforward dimension $64$. Note that in the regular-GPU setting the maximum sequence length resulted in an out of memory error. Furthermore, in the amortized-GPU setting there appears to be no time increase at all. We believe that in the tested regime the forward pass time for amortized-GPU is dominated by overhead and that this is the reason why no time increase can be observed.

\subsection{How does amortized compare to regular inference in a real world setting? }
Finally, we describe our experience deploying concatenating, and appending with and without amortized inference on the LinkedIn Feed and Ads ranking models. Results for latency and CPU usage are shown for both use cases in Table~\ref{tab:feed-results}. We also provide results for Feed's online main engagement metric. As shown, the production latency and CPU usage are significantly reduced when using amortized inference. Furthermore, Feed engagement is higher for amortized inference. This is unexpected given that it is just an inference-optimized version of appending with regular inference. However, online ranking systems are known to show a relationship between latency reductions and engagement increases \cite{kersbergen2021learnings, khandelwal2021jointly}. We thus believe that the further engagement increase can be attributed to the reduced latency in the amortized inference version.

\begin{table}
  \caption{Latency and A/B test results on Feed and Ads.}
  \label{tab:feed-results}
  \begin{tabular}{cccc}
\toprule
\textbf{Metric} & \textbf{Concat} & \textbf{Append} & \begin{tabular}[c]{@{}c@{}}\textbf{Append}\\ \textbf{(amortized)}\end{tabular} \\ \midrule
\multicolumn{4}{c}{\textit{Feed}}                           \\
Latency (p90)   & +52\%           & +56\%           & \textbf{+11\%}                       \\
CPU Usage (p95) & +44\%           & +43\%           & \textbf{+5.5\%}                      \\
Engagement      & +0.14\%         & +0.11\%         & \textbf{+0.18\%}                     \\ \midrule
\multicolumn{4}{c}{\textit{Ads}}                           \\
Latency (p99)   & +86\%           & --           & \textbf{+10\%}                       \\
CPU Usage & +50\%           & --           & \textbf{+10\%}                      \\ \bottomrule
\end{tabular}
\end{table}

\section{Related Work}
The deployment cost of user action history modeling with Transformers is a well known problem. While some works chose to upgrade their infrastructure such as in \cite{xia2023transact}, other works have focused on making inference more efficient.
There has been specific focus on efficiency in the field of lifelong user behavior modeling. Here it is common to have a two stage approach. First, a general search unit (GSU) is used to reduce the size of the user interaction sequence from tens of thousands to tens or hundreds. After that, an exact search unit is applied on the reduced sequence. \cite{pi2020search} uses inner product search for the GSU and then applies multi-head attention on the reduced sequence. \cite{cao2022sampling} improves on this by replacing inner product search with locality sensitive hashing which closely resembles softmax attention. Finally, \cite{chang2023twin} uses exact multi-head attention on the long sequence of interactions, but caches the projection of item features and reduces the projection size of context features such as timestamp and action. While we experimented with an inner product GSU, we observed offline metric drops, likely due to the short term nature of our sequences.

Other works have focused on improving the efficiency of multi-head attention by exploiting behavioral sequence structure. \cite{hu2023ps} observes that attention patterns are sparse and that computation is wasted on items that have low relevance to the candidate. The authors develop a progressive sampling-based self-attention mechanism to identify which items are valuable. \cite{wang2023convformer} replaces multi-head attention with convolution and employs convolution optimizations. Lastly, \cite{zhu2024collaboration} uses multi-query attention to reduce the size and cost of multi-head attention. These approaches are orthogonal to amortized inference and can be combined for further speed ups.

\section{Conclusion}
We have studied user action history encoding for DLRMs with focus on early fusion methods and efficient inference through amortized history inference.  When comparing concatenation and appending of the candidate, we found that there is no one method that consistenly performs better. In particular, on our Feed and Ads offline results the two methods were within the threshold of what is considered significant. This result allows us to choose to append the candidate item to the sequence and use cross-attention which in turn makes it possible to only infer the member history once per request and for all request items at the same time. This amortizes history computation and significantly reduces the computational cost of deploying user action history encoding online on two surfaces at LinkedIn. In online engagement results for Feed we furthermore found append with amortized inference to outperform concatenation and append without amortization which may be due to improved latency.


\bibliographystyle{ACM-Reference-Format}
\bibliography{ms}










\end{document}